%% file: main.tex
\documentclass[10pt, journal]{IEEEtran} 

\IEEEoverridecommandlockouts                              

\usepackage{amsmath,amsfonts}
\usepackage{algorithmic}
\usepackage{algorithm}
\usepackage{array}

\usepackage{textcomp}
\usepackage{stfloats}
\usepackage{url}
\usepackage{verbatim}
\usepackage{graphicx}
\usepackage{cite}

\usepackage{ulem}

\usepackage{times}
\usepackage{siunitx}

\usepackage{multirow}
\usepackage{adjustbox}
\usepackage{booktabs}
\usepackage{makecell}

\usepackage{siunitx}
\DeclareSIUnit[number-unit-product = {\thinspace}]{\inch}{in}

\usepackage[hidelinks]{hyperref}
\hypersetup{colorlinks,breaklinks,
	urlcolor=blue,linkcolor=blue, 
	citecolor=citecol}
\usepackage[dvipsnames]{xcolor}
\definecolor{citecol}{rgb}{0.0,0.6,0.0}

\newcommand{\bs}[1]{\mathbf{#1}}

\begin{document}

\title{Environment-Aware and Human-Cooperative Swing Control for Lower-Limb Prostheses in Diverse Obstacle Scenarios}

\author{Haosen Xing, Haoran Ma, Sijin Zhang, and Hartmut Geyer
\thanks{H.\ Xing, H.\ Ma, S.\ Zhang, and H.\ Geyer are with the Robotics Institute, Carnegie Mellon University, Pittsburgh, PA 15213, USA. {\tt\small \{haosenx,sijinz,hgeyer\}@andrew.cmu.edu, haoram98@gmail.com}.}%
}





\maketitle

\begin{abstract}

Current control strategies for powered lower limb
prostheses often lack awareness of the environment and the user’s
intended interactions with it. This limitation becomes particularly apparent in complex terrains. Obstacle negotiation, a critical scenario exemplifying such challenges, requires both real-time perception of obstacle geometry and responsiveness to user intention about when and where to step over or onto, to dynamically adjust swing trajectories. We propose a novel control strategy that fuses environmental awareness and human cooperativeness: an on-board depth camera detects obstacles ahead of swing phase, prompting an elevated early-swing trajectory to ensure clearance, while late-swing control defers to natural biomechanical cues from the user. This approach enables intuitive stepping strategies without requiring unnatural movement patterns. Experiments with three non-amputee participants demonstrated 100\% success across more than 150 step-overs and 30 step-ons with randomly placed obstacles of varying heights (4-16 cm) and distances (15-70 cm). By effectively addressing obstacle navigation—a gateway challenge for complex terrain mobility—our system demonstrates adaptability to both environmental constraints and user intentions, with promising applications across diverse locomotion scenarios.

\end{abstract}


\input{introduction}

\input{method}

\input{results}
\input{discussion}

\bibliographystyle{IEEEtran}
\bibliography{reference}

\end{document}

%% file: introduction.tex
\section{Introduction}

The ability of lower-limb assistive devices to function effectively is challenged by the unpredictability of real world environments.
Although several approaches to the control of lower limb prostheses and exoskeletons have been developed~\cite{sup2009preliminary, lenzi2014speed, shi2019review, hood2022powered, kim2022seamless, best2023data, gehlhar2023review}, these approaches often struggle to adapt to complex scenarios such as obstacle crossings due to their reliance on pre-programmed motion patterns. Consequently, incorporating environmental information into control has become a promising research direction~\cite{ gionfrida2024wearable} with the potential for lower-limb assistive devices to detect and respond to changes in the environment in real time and therefore to provide safer and more comfortable use~\cite{tschiedel2020relying}.

Research on integrating such environmental information has focused primarily on terrain recognition and classification. This involves using sensors such as laser scanners~\cite{liu2015development, luo2023early}, depth cameras~\cite{krausz2015depth, massalin2017user, zhang2019environmental, karacan2020environment, krausz2021sensor, li2022fusion, ramanathan2022visual, yin2023environmental}, and RGB cameras~\cite{zhong2021efficient, rai2021vision, laschowski2022environment, kurbis2022stair, contreras2023convolutional, tricomi2023environment} to capture and classify the terrain ahead of a user with the goal of adjusting the pre-programmed motion patterns of assistive devices in anticipation of the upcoming environment. However, relying solely on classification has drawbacks. Even with reported classification accuracies ranging from 82\% to 99\%, this would still translate into 3 to 54 falls per month~\cite{tschiedel2020relying}. Furthermore, terrains classified with the same label can vary widely in size and shape, demanding additional refinement in control strategies to effectively mitigate fall risks.

\begin{figure}[t]
\centering
\includegraphics[width=0.5\columnwidth]{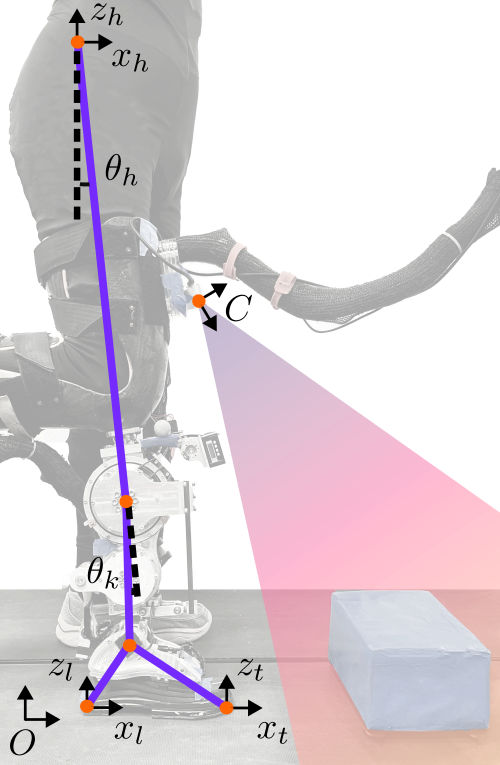}
\caption{Powered knee and ankle prosthesis with depth camera mounted on able-bodied adapter. Kinematic model of user and prosthesis for state estimation and control shown in blue with world ($O$) and camera ($C$) coordinate frames, user hip angle ($\theta_h$, relative to world frame vertical) and position ($x_h$, $z_h$), and prosthesis knee angle ($\theta_k$) and toe ($x_t$, $z_t$) and heel ($x_l$, $z_l$) positions highlighted.}
\label{fig: leg}
\end{figure}

With a focus on lower limb prostheses, recent studies have started to include more detailed information about the environment in the control design. For instance, \cite{cheng2023automatic} and \cite{hong2023vision} incorporate the distance and height of detected obstacles to re-plan the motion of a lower limb prosthesis for the upcoming swing phase. These approaches, however, do not account for the user's intended decision, such as the desire to modify the limb motion \textit{during} the swing phase. Toward a more flexible prosthesis behavior with human intent integrated, \cite{zhang2020subvision} presented a swing controller for stepping over obstacles by continuously modifying the joint velocities of the prosthesis based on its current distance to the obstacle, while the user's thigh motion determined the height of the foot during the step-over. Similarly, \cite{thatte2019real} presented a reactive controller that adapts the motion of the prosthesis throughout swing using estimates of the pose of the prosthesis and of the future path of the user's hip to avoid scuffing the ground and tripping. Nonetheless, both studies predefined the position of the prosthesis at the end of the swing, severely limiting the flexibility of the controller. For example, rather than to step over an obstacle, a user might want to step onto it. An ideal human assistive device should seamlessly integrate such user preferences into the device control \cite{weber2023assistive}.

Here, we present a more human-cooperative prosthesis control that adapts the swing motion of the prosthesis dynamically in response to environmental information and includes the user in making decisions about where and when to land. To this end, we integrate computer vision and state estimation algorithms to create a swing controller that synchronizes the prosthesis's knee motion with the user's estimated hip motion, yet continually adapts this synchronization based on information about obstacles in the path of the prosthesis. By not predefining the final position of the prosthesis, the user maintains authority over where and when landing occurs. We then demonstrate the effectiveness of the proposed control in experiments with three non-amputee participants who use the powered knee and ankle prosthesis with an adapter (Fig.~\ref{fig: leg}) during level walking and when stepping over and onto obstacles of various sizes and distances at the onset of swing. Finally, we discuss the potential of the proposed control to not only allow flexible obstacle negotiation but also generalize to locomotion in any environment.

%% file: method.tex
\section{Swing Control}
\label{s:methods}

\subsection{Environment Information}

\begin{figure*}[t]
\centering
\includegraphics[width=\linewidth]{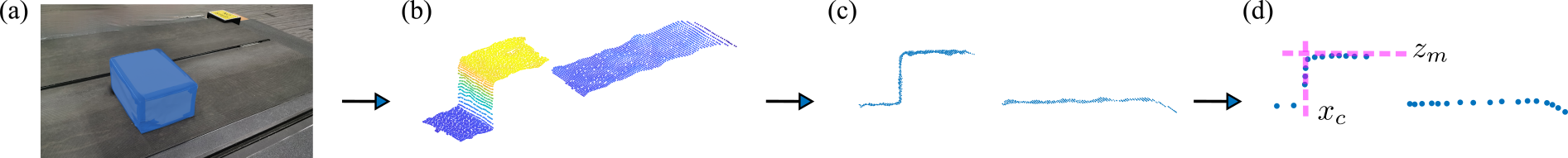}
\caption{Processing steps for extracting environment information. 
(a) Example of obstacle encounter on the treadmill. 
(b) Corresponding 3-D point cloud cropped to a width of \SI{15}{\centi\m} to reduce data processing. 
(c) Point cloud projected onto sagittal plane. 
(d) Pruned elevation map with identified maximum obstacle height ($z_m$) and horizontal distance from the toe location to the (projected) front of the obstacle ($x_c$). }
\label{fig: flowchart}
\end{figure*}

We extract environment information from a depth camera (Intel RealSense D435) mounted on the able-bodied adapter of the knee and ankle prosthesis  (Fig.~\ref{fig: leg}). The camera generates a 3-D point cloud of the terrain ahead of the prosthesis, which we capture during the late stance phase when the camera points downward (Fig.~\ref{fig: flowchart}-a,b). We transform the point cloud into the global coordinate frame $O$ using state estimation based on two IMUs on the prosthesis (mounted above the knee and on the foot) and project it onto the sagittal plane (Fig.~\ref{fig: flowchart}-c). We then prune this projection using K-means clustering~\cite{likas2003global} for key point detection, resulting in an elevation map from which we extract the maximum elevation $z_m^\prime$ and the horizontal distance $x_c$ from the prosthesis toe to the largest elevation change (Fig.~\ref{fig: flowchart}-d). 

For control purposes, we modify the two estimates. First, we modify the elevation estimate to the obstacle height $z_m = \max(z_m^\prime, z_t) + \delta$, where $z_t$ is the prosthesis toe height in late stance and $\delta = \SI{1}{\centi\m}$. Besides introducing a safety margin, this modification ensures that the prosthesis foot lifts off the ground in level walking and when encountering lowering elevation profiles (such as in stair and ramp declines, not investigated here). Second, if indeed $z_m^\prime \leq z_t$, we set $x_c$ to the arbitrary value of \SI{20}{\centi\m}, which leads to a satisfactory lift-off trajectory of the prosthesis in level walking.

\subsection{Prosthesis Knee Swing Controller}
For controlling the prosthesis motion during swing, we assume that the foot remains perpendicular to the shank throughout the swing and model the prosthesis as a two-link manipulator with hip and knee joints ($\theta_h$ and $\theta_k$). The controller uses this model to plan the desired knee velocity of the prosthesis based on the current motion of the user's hip and on the height $z_m$ and distance $x_c$ of the obstacle extracted from the elevation map. More specifically, planning occurs in the joint space, where the current pose of the prosthesis reduces to a point $\bs{P}(\theta_h, \theta_k)$ and obstacle information defines forbidden areas that change over time~\cite{warren1989global}. The planning uses estimates of the hip, toe and heel locations of the prosthesis ($O$-frame coordinates $x_h$, $z_h$, $x_t$, $z_t$, $x_l$ and $z_l$) and of the angles and angular velocities of the hip and knee ($\dot{\theta}_h$, $\dot{\theta}_h$) and the knee angular acceleration ($\ddot{\theta}_k$) (details on state estimation in a recent work [citation omitted for anonymity]) (Fig.~\ref{fig: leg}). Planning is divided into three phases.  

\subsubsection{Phase One--Raise Toe Above Obstacle Height}
The goal of phase one is to raise the prosthesis toe above the obstacle height $z_m$ before $x_t$ reaches the location $x_c$. To achieve this goal, we consider two regions of the joint space shown in figure~\ref{fig: phases}-b. The first region, $M_x^t$, marks all joint configurations in which the toe has not yet reached the obstacle, $x_t < x_c$ (light blue). The second, $M_z^t$, represents joint configurations with the toe below the estimated obstacle height, $z_t < z_m$ (red). The superscript $t$ indicates that these regions change dynamically; for example, as the user moves forward and elevates the hip, $M_x^t$ shifts leftward and $M_z^t$ shrinks. Due to the shift to the left, the joint configuration $\bs{P}^t$ of the prosthesis will exit $M_x^t$ eventually, and successful control in phase one requires $\bs{P}^t$ to exit $M_z^t$ beforehand, as the latter becomes a forbidden region once $x_t \geq x_c$. Given an initial configuration $\bs{P}^0$, the correct exit order can be achieved by moving along the slope $k_1^t = \frac{\Delta\theta_k^t}{\Delta\theta_h^t}$, where $\Delta \theta_k^t$ is the vertical distance to the edge of $M_z^t$ and $\Delta \theta_h^t$ is the horizontal distance to the edge of $M_x^t$. Thus, in phase one, we set the desired knee angular velocity $\dot{\theta}_k^{t+1}$ of the prosthesis for the next time step ($t+1$) to
\begin{align} \label{e:1}
    \dot{\theta}_k^{t+1} = k_1^t \dot{\theta}_h^t \,,
\end{align}
where $\dot{\theta}_h^t$ is the estimated hip angular velocity of the user. Note that if an obstacle detected by the camera is very far away, the region $M_x^t$ may expand far to the right of the configuration space and the vertical distance $\Delta\theta_k^t$ cannot be established. To ensure fast enough lift off in this case, we enforce a lower threshold $k_1^t \geq k_{min}^t$, where $k_{min}^t$ corresponds to the slope obtained assuming that $\Delta\theta_k^t$ is the vertical distance between the current point $\bs{P}^t$ and the peak of the region $M_z^t$.

\subsubsection{Phase Two--Maintain Toe Height}
\begin{figure*}[t]
\vspace{0.2cm}
\centering
\includegraphics[width=\linewidth]{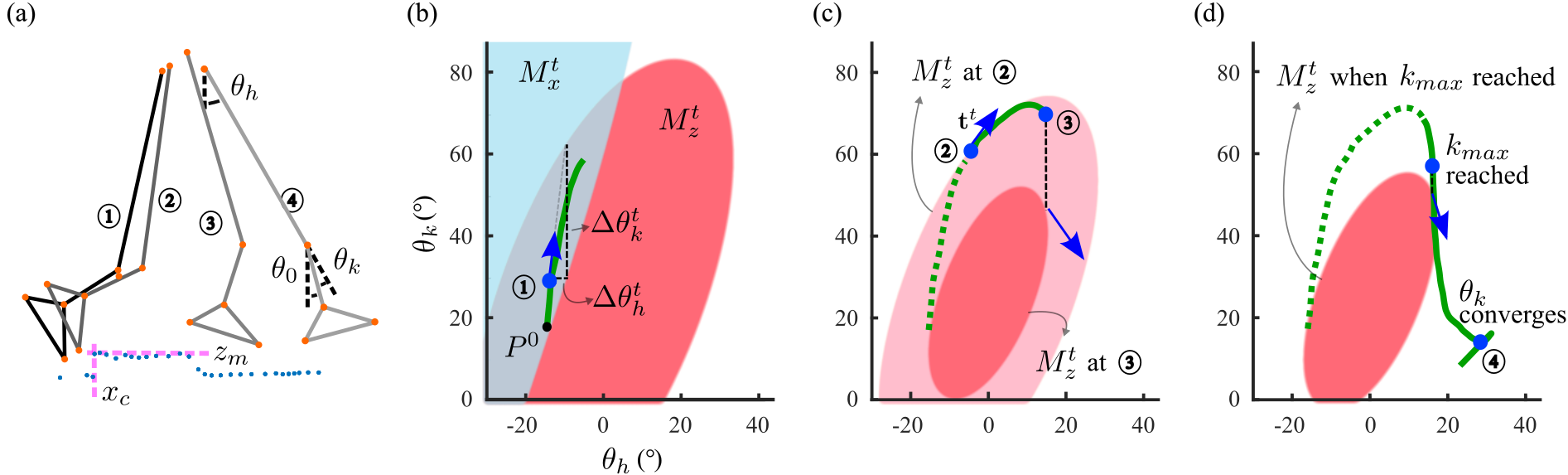}
\caption{Swing control overview. (a) Elevation map and kinematic model at start of swing (1), transitions to phase two (2) and three (3), and when the knee angle starts to mirror the hip angle in phase three (4). 
(b) Phase one with example of current regions of joint
configurations for which prosthesis toe has not yet reached obstacle ($\bs{M}_\bs{x}^t$, light blue) and for which the toe is below the obstacle height ($\bs{M}_\bs{z}^t$, red). Also shown are generated trajectory from initial configuration $P^0$ with configuration corresponding to kinematic model (1) marked as blue point. Dashed lines indicate horizontal and vertical distances of hip ($\Delta\theta_h^t$) and knee angle ($\Delta\theta_k^t$) to region boundaries
(c) Phase two with forbidden region $M_z^t$ at beginning (light red) and end (red) of the phase. Blue arrows show respective tangents $\bs{t}^t$ of region boundary at current hip angle. (d) Phase three with forbidden region at moment when magnitude of slope reaches $k_{max}$ (compare text) and trajectory of $\bs{P}^t$ until the touchdown. Once the knee angle has converged to $\theta_k^t = \theta_h^t - \theta_0$ (4), changes in the knee angle strictly mirror changes in the hip angle and the prosthesis shank maintains a constant forward lean $\theta_0$ in preparation for landing (see panel (a)).}
\label{fig: phases}
\end{figure*}

Once the toe height of the prosthesis reaches the obstacle height, the control switches to phase two with the goal of maintaining $z_t$ above $z_m$. One way to achieve this goal is to strictly follow the contour of the forbidden region $M_z^t$ in the joint space. However, such a strict policy leads to unnecessarily aggressive prosthesis behavior, as $M_z^t$ tends to shrink rapidly due to hip elevation by the human user. Instead, we rely on the tangent vector $\bs{t}^t$ of the contour (Fig.~\ref{fig: phases}-c), which varies much less rapidly and generates a smoother prosthesis behavior. Specifically, we set the desired knee angular velocity to the instantaneous slope $k_2^t$ of the tangent vector,
\begin{align}
    \dot{\theta}_k^{t+1} = k_2^t\dot{\theta}_h^t. \label{eq:2}
\end{align}
Note that when $k_2^t < 0$, $M_z^t$ remains unchanged to ensure further smoothness and prevent potential $M_z^t$ inflation.

\subsubsection{Phase Three--Prepare For Landing}
Once the horizontal position of the heel passes that of the hip, $x_l > x_h$, the control switches to phase three. In this phase, the swing leg points forward, and the goal is to extend the knee and prepare the leg for landing regardless of whether the user intends to step onto or over the obstacle. The extension happens largely automatically by continuing to set the desired angular velocity of the knee according to \eqref{eq:2}, as the slope of the tangent of $M_z^t$ is negative when the heel has passed the hip position (compare Fig.~\ref{fig: phases}-c). However, we introduce a few practical tweaks. First, since the hip velocity can sometimes go to zero, we use its iterative maximum, $\dot{\theta}_h^t = \max_{\tau}\, \dot{\theta}_h^\tau$ to prevent freezing of the knee, where $\tau$ goes from the start time of phase three to the current time. Second, because the tangent will inevitably point straight down at some point in this phase (Fig.~\ref{fig: phases}-c), we prevent excessive knee velocities by saturating the magnitude of the slope to $k_{max}$. Third, once the slope reaches this saturation level, we switch the control to  
\begin{align}
    \dot{\theta}_k^{t+1} = C^t\, k_{max} \dot{\theta}_h^t
    \label{eq:3}
\end{align}
with 
\begin{align}
    C^t = \frac{\theta_k^t - \theta_h^t + \theta_0}{\theta_k^* - \theta_h^* + \theta_0}\,, 
\end{align}
where the term $C^t$ ensures the knee converges to an angle $\theta_k^t = \theta_h^t - \theta_0$ that is offset from the hip angle by $\theta_0$, and $\theta_k^*$ and $\theta_h^*$ are the knee and hip angles at the time when the slope reached the saturation level. Finally, once the knee angle has converged, we apply 
\begin{align}
    \dot{\theta}_k^{t+1} = \dot{\theta}_h^t\,, \label{eq:4}
\end{align}
and the prosthesis shank maintains the same pose (forward lean by offset angle $\theta_0$ with respect to the world frame vertical) in preparation for landing until the prosthesis foot detects contact. (The parameters $\theta_0$ and $k_{max}$ are hand-tuned.)

\begin{figure}
\centering
\includegraphics[width=\linewidth]{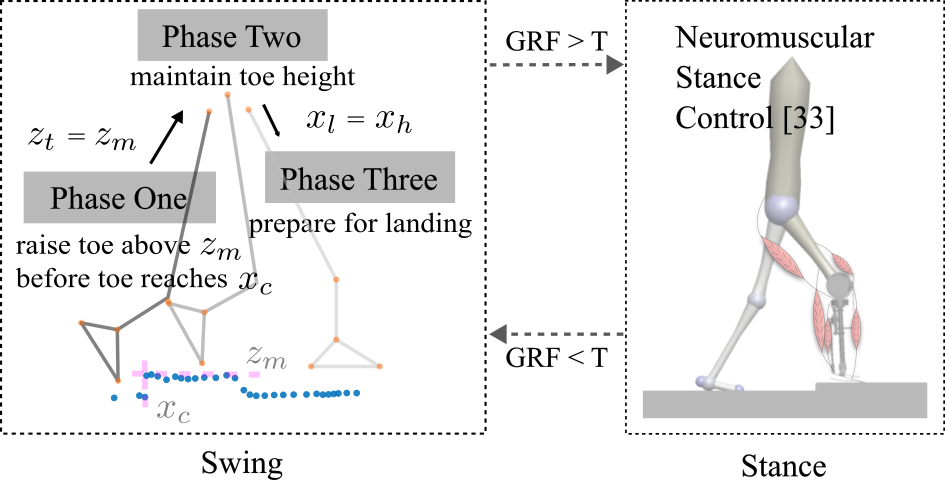}
\caption{Prosthesis controller overview. The controller is either in stance or swing phase, depending on the prosthesis leg ground reaction force crossing a threshold $T$. Stance control uses a virtual neuromuscular model mimicking human leg behavior \cite{thatte2015toward}. Swing control is implemented as detailed in section~\ref{s:methods} and investigated in this paper.}
\label{fig: sm}
\end{figure}

\subsubsection{Phase transitions}
To ensure smooth prosthesis motions, the velocity command send to the prosthesis knee gradually transitions at the beginning of a new phase,
\begin{align} \label{e:knee_vel_command_swing}
    \dot{\theta}_{k,cmd}^{t+1} = \left(1-\gamma_1\right) \dot{\theta}_k^{t+1} + \gamma_1\left(\dot{\theta}_{k,msd}^{t} + \gamma_2\ddot{\theta}_{k,ini}\,\Delta t \right)\,,
\end{align}
where $\dot{\theta}_{k,msd}^{t}$ is the measured prosthesis knee velocity, $\ddot{\theta}_{k,ini}$ is the knee acceleration at the start of the new phase, $\Delta t$ is the controller time step, and $\gamma_i = e^{-\alpha_i n}$ are exponential decay terms defined by tuning parameters $\alpha_i$ and the number $n$ of time steps since the start of the new phase.

%% file: results.tex
\section{Prosthesis Experiments and Results}

\begin{figure*}
\centering
\includegraphics[width=0.95\linewidth]{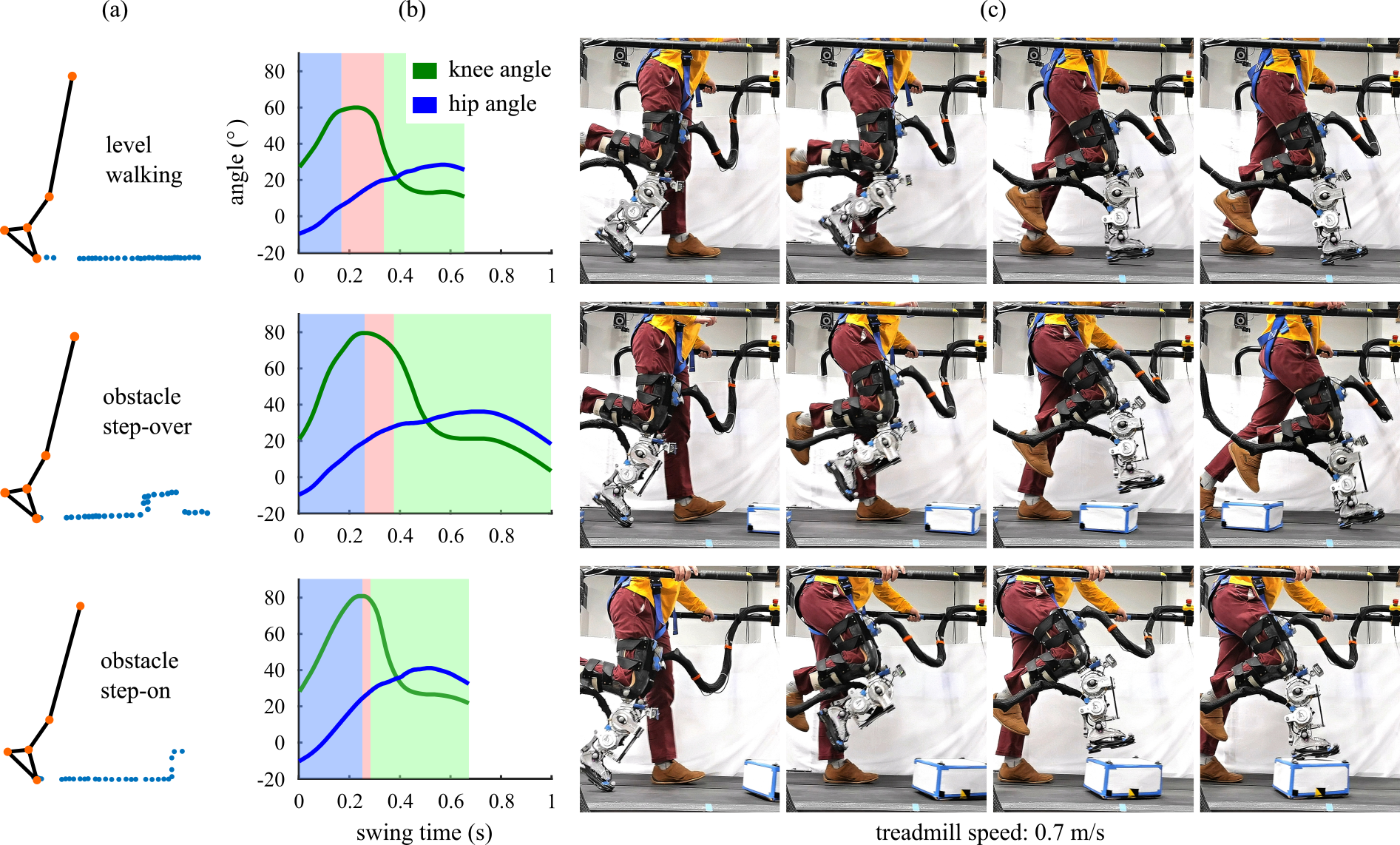}
\caption{Representative examples for experimental results in level walking (top), stepping over (middle) and stepping onto (bottom) obstacles. (a) Elevation maps at time of prosthesis leg toe-off. (b) Estimated swing trajectories of user hip angle $\theta_h$ (blue) and resulting prosthesis knee angle $\theta_k$ (green). Swing control phases (see Sec.~\ref{s:methods}) shaded in light blue, pink, and green for phases one, two and three, respectively. (c) Video capture still frames at toe-off, the transition between phases one and two, the moment the knee angle converges to $\theta_k^t = \theta_h^t - \theta_0$, and at heel-strike.} 
\label{fig: experiment}
\end{figure*}

\subsection{Prosthesis Control Implementation}
The controller of the active knee and ankle prosthesis is implemented within the Simulink Real-Time environment and operates at a \SI{1}{\kilo\hertz} update rate. The prosthesis knee and ankle joints are driven by two separate series elastic actuators, which can be operated in torque or position control. Two IMUs mounted above the knee and on the foot of the prosthesis are used for state estimation. During swing, the prosthesis knee is controlled as detailed in section~\ref{s:methods} (Fig.~\ref{fig: sm}, left panel). Specifically, the resulting knee velocity command (\ref{e:knee_vel_command_swing}) is integrated and tracked with position control \cite{thatte2019robust}. The prosthesis ankle joint follows a minimum jerk trajectory that is generated at toe-off and returns the ankle to a perpendicular position. Although the focus of this paper is on swing control, the prosthesis implements a virtual neuromuscular control for both the knee and ankle during stance that mimics human leg behavior in this phase~\cite{thatte2015toward} (Fig.~\ref{fig: sm}, right panel).

\subsection{Experiments}

We evaluated the effectiveness of the proposed swing control through treadmill locomotion experiments with three non-amputee participants (one female, two male) (Fig.~\ref{fig: experiment}). All participants provided written informed consent as per a protocol approved by the 
University Review Board. The participants used the knee and ankle prosthesis with an able-bodied adapter and engaged in level walking as well as stepping over and onto obstacles while the treadmill operated at a speed of \SI{0.7}{\m\per\s} for safety. The obstacles consisted of four cardboard boxes: three with heights of \SI{4}{\centi\m}, \SI{8}{\centi\m}, and \SI{16}{\centi\m} for step-overs and one reinforced \SI{16}{\centi\m} box for step-ons. During a trial, we randomly selected these four boxes 15 to 20 times and placed them on the belt at the front of the treadmill.  Apart from the goal of stepping over or onto the obstacles, we instructed the participants to lightly hold onto the handrails for safety but only support their bodies through the handrails when they felt uncomfortable. We gave no additional guidance. All participants had ample practice walking with the prosthesis, and we conducted the experiments only once they reported feeling comfortable. We recorded at least three trials per participant and task (level walking and obstacle encounters), which resulted in observing at least 100 walking steps, 50 step-overs, and 10 step-ons per participant. For observing ground truth, we used motion capture at \SI{200}{Hz} of reflective markers placed on the great trochanter of the participant and the knee, toe and heel of the prosthesis.

\subsection{Results}

Across the many steps of walking on level ground and over and onto obstacles of various heights and initial distances from the prosthesis, the control maintained a 100\% success rate for all three participants, as long as the camera had the obstacle in its view at the time of take-off. On rare occasions, an obstacle was placed too far away and the control incorrectly assumed level ground, causing the prosthesis toe to strike the obstacle late in the swing phase. Rather than the result of the overall control strategy, these failure instances arose from a specific limitation of the depth camera we used. The camera's field of view is \SI{65}{\degree}, which limited the distance it could look ahead on the ground to \SI{1}{\m}. This distance was not enough to cover the largest initial distances to the obstacles that occurred.

\begin{figure*}
\centering
\includegraphics[width=\linewidth]{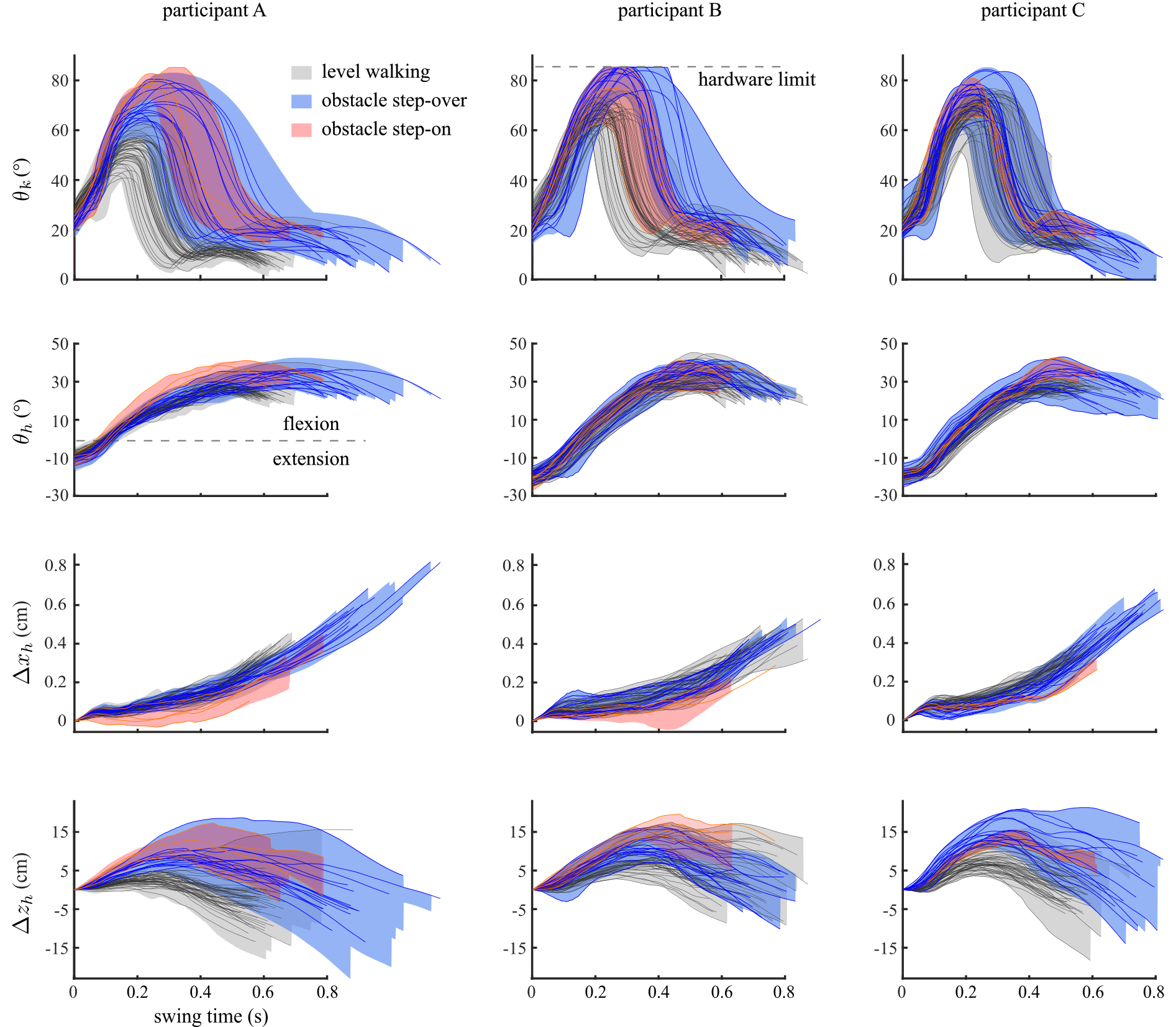}
\caption{Swing-phase kinematic trajectories of the three participants for all trials.
(Top row) Prosthesis knee angle, (second row) user’s hip angle, (third row) hip forward position (relative to take-off), and (bottom row) hip height (relative to take-off) for each participant (columns). Shaded areas indicate the range of trajectories for level walking (gray), and stepping over (blue) and onto obstacles (red), with the areas' upper and lower bounds marking the largest and smallest observed values. The upper bound of the prosthesis knee angle is hardware-limited to \SI{85}{\degree}. Thin solid lines show randomly selected example trajectories.}
\label{fig: subjects}
\end{figure*}

\subsubsection{Environment Awareness}
Environment awareness was critical to raise the foot for obstacle negotiation. Figure~\ref{fig: experiment}-b shows three representative examples of the user hip and prosthesis knee trajectories during level walking (top), obstacle step-over (middle) and step-on (bottom). In all three cases, the user hip performed similarly (blue trajectories) in the first \SI{300}{\milli\s} of swing, suggesting the participants did not actively intervene when the swing control was in phase one and raised the prosthesis foot above the obstacle height. In contrast, to achieve foot clearance, the prosthesis knee flexed significantly more with obstacles present (reaching about \SI{80}{\degree} peak flexion \textit{vs.} \SI{60}{\degree} in level walking). This increased knee flexion is also visible in the video capture still frames at the transition between phases one and two (Fig.~\ref{fig: experiment}-c), where the prosthesis foot is clearly above the box as intended. 

The findings from the representative examples generalize to all three participants and trials. Figure~\ref{fig: subjects} summarizes the swing trajectories of the prosthesis knee angle and the user hip angle and height observed in the three tasks for all three participants. Again, the user hip angle trajectory remains similar for all three tasks in the first \SI{300}{\milli\s} of swing ($\theta_h$, middle row), while the prosthesis knee flexes substantially more during obstacle encounters ($\theta_k$, top row). In addition, while subjects A and C elevated their hips in the obstacle tasks more than in the level walking  ($\Delta z_h$, bottom row), the similarity of the hip height changes between the three tasks for subject B show that this active intervention was not critical. 

\subsubsection{Human Cooperativeness}
The swing control was effective at including the participants in determining the prosthesis motion during swing. By control design, a user can influence the prosthesis swing motion in three essential ways: speed up or slow down the entire swing as the prosthesis knee motion is synchronized to the user hip angular velocity throughout (Eqs.~\ref{e:1}-\ref{eq:4}); lengthen or shorten the prosthesis step by progressing more or less forward on the contralateral stance leg, as phase two (maintain toe height) of the control only transitions to phase three (prepare for landing) once the horizontal position of the prosthesis heel catches up with the user hip ($x_l = x_h$); and effect landing of the prosthesis by extending and lowering the hip in phase three, as the control only locks in a constant forward lean $\theta_0$ of the prosthesis shank without predefining a landing point. The participants used two of these three ways to negotiate the obstacles. They used less hip forward progression (Fig.~\ref{fig: experiment}-b, early termination of control phase 2, and Fig.~\ref{fig: subjects}, hip forward position $\Delta x_h$) to effect shorter steps onto the obstacles and more forward progression combined with delayed landing in phase three to effect larger steps over obstacles (Fig.~\ref{fig: experiment}-c, last two still frames, and Fig.~\ref{fig: subjects}, average swing time \SI{0.81}{\s} for step-overs \textit{vs.} \SI{0.64}{\s} for step-ons and \SI{0.61}{\s} for level walking). The participants did not significantly speed up or slow down swing through faster or slower hip angular motion (Fig.~\ref{fig: subjects}, middle row), which may be due to the fixed treadmill speed used in the experiments. Overall, the indirect influence on the prosthesis control felt natural to the participants and proved effective at incorporating them in determining the swing motion of the prosthesis.

\begin{figure*}
\centering
\includegraphics[width=\linewidth]{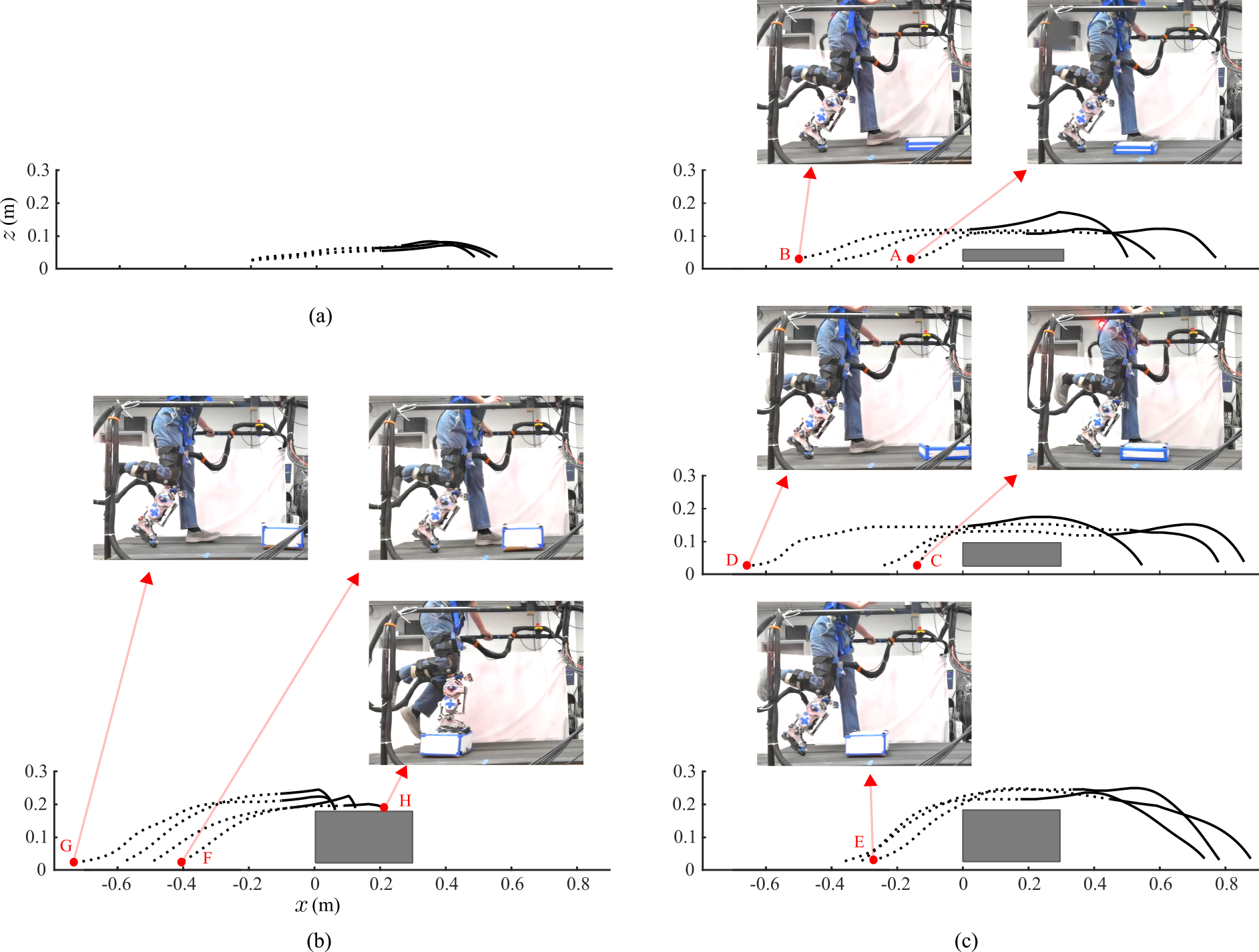}
\caption{Example foot height trajectories for participant A in level walking (a), and stepping on (b) and over (c) obstacles. Foot height defined as the lower value of toe and heel heights captured by the motion capture system. The solid line portions mark part of trajectories where swing control applies equation~\ref{eq:4} and prosthesis shank maintains constant forward lean $\theta_0$. Obstacles indicated as gray boxes and x-axis defined relative to point $x_c$ identified from the environment map.
Video capture still frames from selected toe-off moments (A-G) and moment after stepping on the obstacle (H) illustrate the variability in obstacle height and initial distance as well as in task that the proposed swing control successfully handles.} 
\label{fig: vicon}
\end{figure*} 

To further demonstrate the human-cooperativeness of the proposed swing control, figure~\ref{fig: vicon} shows for one participant the foot height trajectories of representative steps during level walking and obstacle encounters. The trajectories demonstrate that the control is effective at providing sufficient foot clearance for various obstacle heights (\SI{0}{\centi\m} to \SI{16}{\centi\m}), allows the participant to negotiate the obstacles from varying distances at the onset of swing (\SI{70}{\centi\m} to \SI{15}{\centi\m}), and enables the participant to effect lowering and thereby determine the landing location for a significant portion of the mid to late swing (Eq.~\ref{eq:4} active, solid line portion of trajectories).

%% file: discussion.tex
\section{Discussion}

We presented a novel prosthesis control that adapts the swing motion of a powered lower limb prosthesis to an obstacle in the path while including the human user in determining where and when to land. In validation experiments with three non-amputee participants, we demonstrated the effectiveness of the control when negotiating obstacles of varying sizes and initial distances to the prosthesis as well as when pursuing the competing goals of stepping over or onto the obstacles. To our knowledge, this work is the first to demonstrate real-time control of a lower limb prosthesis that integrates both environment information and user interaction.

Our findings support the claim that environment information is vital for broadening the locomotion activities provided by wearable robots and enhancing their quality of assistance~\cite{gionfrida2024wearable}. Without environment information, lower-limb prostheses rely on specific, unnatural motion cues to alter swing and navigate more complex landscapes. For example, commercial knee prostheses like the OttoBock Genium series~\cite{bellmann2012stair} require the users to extend their hip backwards (contrary to the natural flexion forward) at the beginning of swing to trigger a passive yet rapid flex and extend motion of the prosthesis knee that enables stair climbing and stepping on and over obstacles. Similarly, control strategies proposed in the research community for powered lower-limb prostheses require unnatural hip extension at the onset of swing to either lock the prosthesis knee \cite{rezazadeh2019phase} or flex it in proportion to the duration of the backward extension \cite{mendez2020powered} to step over obstacles. While effective to an extent, it is unclear how such motion cues may generalize to a wider array of locomotion tasks, such as climbing ladders or rocks. None of these complications seem necessary if environment information becomes integral to prosthesis control. Yet, control strategies that solely rely on environment information \cite{cheng2023automatic,hong2023vision,zhang2020subvision,thatte2019real} deprive users of agency and, thus, equally restrict the locomotion tasks that can be addressed. On the other hand, our proposed control strategy leverages environment information to adapt the prosthesis motion to obstacles from early to mid swing while taking advantage of biomechanical cues that naturally occur from mid to late swing (transition from backward to forward lean of the prosthetic leg, lowering and extension of the hip at the end of swing) to let the user shape the prosthesis motion, suggesting how environment information and human interaction may be combined synergistically to enhance the quality of assistance. 

Further research will be needed to fully evaluate the potential of the proposed control strategy. First, it has yet to be tested with amputee users. During swing, transfemoral amputees show a reduced hip range of motion in the sagittal plane and hip hiking in the frontal plane when compared to able-bodied walking~\cite{sagawa2011biomechanics}. The first difference occurs due to ischium-socket interference~\cite{rabuffetti2005trans} and affects hip extension in the early phase of swing while the second difference is believed to compensate for the limited dorsiflexion in commercial prosthetic ankles~\cite{su2007gait}. Neither of these differences should have a significant impact on the proposed control, as it adapts the prosthesis motion to changes in the user hip angle (Eqs.~\ref{e:1}-\ref{eq:4}) and hip translational motion (region $M_z^t$ changes with hip height and forward progression). However, amputee users may not feel safe using such an adaptive prosthesis control and reject it.

Second, we only demonstrated obstacle negotiation in a confined treadmill locomotion environment, and it remains open whether the control strategy generalizes to other locomotion tasks and realistic environments. Although designed for stepping over and onto obstacles, extending the control to other common locomotion tasks such as negotiating ramps and stairs should be possible, as going up a ramp or stairs resembles stepping onto an obstacle repeatedly and the swing motion during downward steps may be likened to the second half of the swing motion when stepping over an obstacle. In fact, in preliminary tests we found that the current swing control worked equally well for stepping down from an obstacle as for stepping onto and over it (see supplementary video: \href{https://www.youtube.com/watch?v=EhQMb784INM}{movie~S1}). On the other hand, the current information we extract from the environment (obstacle height and distance) is clearly insufficient in more varied and realistic environments. For instance, a flight of stairs will have multiple elevations and a human user may want to take one, two or more steps at once. An exciting direction for future research would be to combine our work with work on smart glasses for environment identification~\cite{rossos2024ai} and gaze detection~\cite{matthis2018gaze}. For instance, humans locate future, intended foot placements with a constant look-ahead time~\cite{matthis2018gaze}, which could be used to identify which step of a flight of stairs a user wants to use. In addition, such a transition to smart glasses for extracting environment information (and intent) could also alleviate concerns about the practicality of a prosthesis-bound vision system.
